\RequirePackage{fix-cm}
\documentclass{fairmeta}

\DeclareFontShape{T1}{optimistic}{b}{n}{<-> s * [0.88] assets/optimistic}{}
\DeclareFontShape{T1}{optimistic}{bx}{n}{<-> s * [0.88] assets/optimistic}{}
\DeclareFontShape{T1}{optimistic}{b}{sc}{<-> s * [0.88] assets/optimistic}{}
\usepackage{amsmath,amsfonts,bm}

\def\eqref#1{equation~\ref{#1}}

\def\1{\bm{1}}

\DeclareMathAlphabet{\mathsfit}{\encodingdefault}{\sfdefault}{m}{sl}
\SetMathAlphabet{\mathsfit}{bold}{\encodingdefault}{\sfdefault}{bx}{n}

\usepackage{amssymb}
\usepackage{float}
\usepackage{algorithm}
\usepackage{algpseudocode}
\usepackage{capt-of}
\usepackage{wrapfig}
\usepackage{needspace}
\usepackage{pgfplots}
\usepackage{pgfplotstable}
\usepgfplotslibrary{fillbetween}
\pgfplotsset{compat=1.18}
\usepackage{textcomp}
\usepackage{listings}
\usepackage{url}

\lstdefinestyle{generationprompt}{
    basicstyle=\ttfamily\fontsize{8}{10}\selectfont,
    columns=fullflexible,
    keepspaces=true,
    breaklines=true,
    breakatwhitespace=true,
    showstringspaces=false,
    upquote=true,
    numbers=none,
    frame=single,
    framerule=0.4pt,
    rulecolor=\color{black!25},
    backgroundcolor=\color{black!3},
    framesep=6pt,
    xleftmargin=6pt,
    xrightmargin=6pt,
    aboveskip=8pt,
    belowskip=8pt
}

\newcommand{\method}{\textsc{FROST}}
\DeclareRobustCommand{\circled}[1]{\textcircled{\scriptsize #1}}

\title{Let Training Guide Selection: Online Synthetic Data Filtering via Real-Anchored Utility}

\author[1,*]{Yanran Wu}
\author[2]{Sana Lakdawala}
\author[2]{Renzo Tassara Miller}
\author[2]{Chongyang Bai}
\author[2]{Sharath Ciddu}
\author[2]{Shivendra Pratap Singh}
\author[2]{Kungang Li}
\author[2]{Sandeep Pandey}
\author[1]{Chunwei Liu}

\affiliation[1]{Purdue University}
\affiliation[2]{Meta}
\contribution[*]{Work done at Meta.}
\date{September 23, 2026}

\abstract{Synthetic data can scale training supervision when real-world data are
limited, but noise and distribution mismatch can reduce its value.
Existing synthetic data selection methods often emphasize fidelity or diversity 
rather than the learner's evolving needs. We propose \method{}, an online
framework that estimates synthetic-data utility through gradient feedback
anchored in real training data. It calibrates batch utility against recent
history to determine \emph{when} filtering is needed and filters samples
only in out-of-band batches to determine \emph{what} to retain, without an
external verifier or held-out validation set. 
 Experiments on two public benchmarks for image classification and LLM
  fine-tuning for text-to-SQL show that \method{} filters out around
  $20$--$30\%$ of the synthetic data while improving
  real-task performance compared with training on the full synthetic data
  pool. We further apply \method{} during training in a large-scale industrial ads
  re-ranking system, achieving significant performance gains over a
  highly optimized production baseline, demonstrating its effectiveness
  and generalizability.
}
\hypersetup{
    pdftitle={Let Training Guide Selection: Online Synthetic Data Filtering via Real-Anchored Utility},
    pdfauthor={Yanran Wu, Sana Lakdawala, Renzo Tassara Miller, Chongyang Bai, Sharath Ciddu, Shivendra Pratap Singh, Kungang Li, Sandeep Pandey, Chunwei Liu}
}

\begin{document}
\maketitle

\section{Introduction}
\label{sec:introduction}

As modern deep learning models scale, their growing demand for training data
makes the availability of real-world data an increasingly important
bottleneck \citep{villalobos2024data}. Synthetic data provides a scalable
alternative when collecting or annotating additional real examples is
costly or constrained. In computer vision, generated images augment
supervision when data access or expert annotation is limited
\citep{fridadar2018gan}; in large language models (LLMs), synthetic text
and instruction--response pairs support pretraining and instruction tuning
\citep{gunasekar2023textbooks,wang2023selfinstruct}. Synthetic data is also
increasingly used in recommendation systems, where sparse interactions,
incomplete user information, and privacy constraints limit available
supervision. For example, LLMRec and DALLRec use LLMs to infer additional
user--item interactions and enrich user or item descriptions for
recommender training \citep{wei2024llmrec,mao2025dallrec}. SCALR generates
synthetic target-domain interactions from observed user behavior in other
domains and combines them with real interactions to train downstream
recommenders \citep{wang2026crossdomain}.

Despite advances in generative quality, synthetic data can still contain
semantic errors and artifacts, miss important variations, or deviate from
the target distribution \citep{geng2024unmet,adamkiewicz2026pretty}. More
importantly, these failures are not always identifiable from individual
samples alone. In image recognition, synthetic data can scale less
effectively than real data \citep{fan2024scaling}, and even images from
more advanced generators can yield lower real-test classification accuracy
due to distributional mismatch with real data
\citep{adamkiewicz2026pretty}. In recommendation systems, generated
user--item interactions and attributes can likewise introduce uncertain
or noisy supervision, since synthetic interactions often lack real user
feedback for verification
\citep{wei2024llmrec,mao2025dallrec,wang2026crossdomain}. This raises a central
challenge: \emph{how can we select synthetic data that provides useful
supervision and improves model performance on real data?}

Existing synthetic-data selection methods often emphasize \emph{fidelity}
and \emph{diversity} of the synthetic data. They use pretrained feature
extractors to assess image--label consistency \citep{he2023synthetic},
select from feature clusters that are close to real examples, or match synthetic
and real feature statistics
\mbox{\citep{hulkund2025datas3,rezaei2026highdimensional}}.
These criteria characterize properties of the data, but do not track
the evolving needs of the model being trained. Moreover, extracting
features and preprocessing the full candidate pool can also be resource-intensive at
production scale. Instead, we estimate synthetic-data utility from training dynamics
by aligning each synthetic sample's gradient with a smoothed
real-training gradient reference. This provides a local signal
of whether a synthetic update supports the real-data objective.
Existing methods also employ training dynamics, but for data-efficient
subset selection on real training datasets
\citep{qin2024infobatch,jin2026orderdp}, with some using held-out target
examples as the selection reference
\citep{xia2024less,wang2024greats}.

In this paper, we introduce \method{}
(\underline{\textbf{F}}iltering with \underline{\textbf{R}}eal data for
\underline{\textbf{O}}nline \underline{\textbf{S}}ynthetic-data
\underline{\textbf{T}}raining),
an online synthetic-data filtering framework built from two key ideas:
(1) estimating synthetic-data utility through gradient feedback anchored
in real training data, and (2) conditioning sample-level filtering on
batch-level utility. For utility estimation, we use gradients from the
model being trained, without a separate feature extractor or offline
preprocessing of the synthetic pool for selection. Specifically, we
maintain an exponential moving average of gradients from real training
batches and score each synthetic sample by its gradient alignment with
this reference. Individual utility estimates are local and affected by
stochastic gradient noise \citep{faghri2020gradientvariance}, and
filtering every batch based on these estimates can discard useful
supervision. We therefore aggregate sample utilities and calibrate the
batch mean against recent training history to determine \emph{when}
sample-level filtering is needed. We retain
batches within a moderate utility band and apply sample-level
filtering using interquartile-range (IQR) thresholds only to batches
outside the band. Sample-level filtering can thus retain useful examples from batches
  that would otherwise be rejected entirely.

 We first evaluate \method{} on two public benchmarks for image
  classification and LLM fine-tuning for text-to-SQL, comparing against
  existing selection methods at matched synthetic-data drop ratios. 
    \method{} outperforms these baselines and improves absolute accuracy
  by $0.58$--$1.14\%$ across 3 synthetic image generators and $0.9\%$
  on text-to-SQL over training with the full synthetic pools,
  but with $20$--$30\%$ fewer synthetic training samples.
  We further apply \method{} to the training of a large-scale industrial ads re-ranking
  system, where it turns a $0.211\%$ Normalized Entropy (NE) regression
  from training with the full noisy synthetic pool into a $0.096\%$
  improvement, compared to real-only training reference.
  Our $0.096\%$ relative NE improvement is well above the 
  $0.02\%$ gain which is considered significant in prior studies
  of well-optimized production recommenders
  \citep{li2022frequencyaware,lai2023adaembed}.
Our contributions are threefold:
\begin{itemize}
\item We propose \method{}, a model- and task-agnostic framework
for online synthetic-data filtering that estimates utility through
gradient feedback anchored in real training data and uses conditional
batch-to-sample filtering to select useful synthetic supervision.

\item We demonstrate the effectiveness and generalizability of
\method{} across image classification and LLM fine-tuning, including
3 synthetic image generators, improving real-task performance with
fewer synthetic training samples.

 \item We achieve significant performance gains over a highly optimized
  production baseline in a large-scale industrial ads re-ranking system,
  highlighting the practical impact of \method{}.

\end{itemize}

\section{Related Work}
\label{sec:related-work}

\noindent\textbf{Synthetic data for deep learning.}
Synthetic data has been explored across many deep learning tasks to enrich
training supervision when real data are scarce or costly to annotate.
In computer vision, text-to-image diffusion models turn class-specific text descriptions into training images \citep{rombach2022ldm,he2023synthetic}.
\citet{shipard2023diversity} improve generation diversity to train
classifiers without real images, while \citet{azizi2023synthetic} use
generated images to augment real-data training. For LLMs, synthetic text
and instruction--response pairs support pretraining and instruction tuning
with less reliance on human-written examples
\citep{gunasekar2023textbooks,wang2023selfinstruct}.
OmniSQL introduces SynSQL-2.5M by automatically generating diverse databases,
questions, and SQL queries, expanding text-to-SQL training coverage without
large-scale manual annotation \citep{li2025omnisql}. In recommendation,
the motivation is to enrich sparse interactions and incomplete side
information. LLMRec and DALLRec generate additional interactions and item
information \citep{wei2024llmrec,mao2025dallrec}, while SCALR transfers
cross-domain behavior into synthetic target-domain events
\citep{wang2026crossdomain}. These synthetic data generation
approaches expand the available supervision;
our goal is to select useful synthetic data after generation.

\noindent\textbf{Synthetic data selection.}
Existing synthetic-data selection methods focus on fidelity and
diversity through semantic filtering, feature-space coverage, or
distribution matching. Pretrained feature extractors are used to assess
similarity to class descriptions or real examples
\citep{he2023synthetic,lin2023explore}.
DS3 selects from synthetic feature clusters close to real examples,
while CovMatch greedily matches the selected subset's feature covariance
to that of real data
\citep{hulkund2025datas3,rezaei2026highdimensional}.
These methods typically select data before downstream training,
requiring feature extraction and preprocessing of the candidate pool.
More fundamentally, as the model evolves during training,
the usefulness of synthetic data can change, while their offline feature-based selection scores remain fixed.
\method{} instead estimates utility online through gradient alignment
with real training data, providing a local estimate of whether a
synthetic update would reduce the current real-data loss.

\noindent\textbf{Data selection based on training dynamics.}
Training-dynamics-based methods use losses and gradients to identify
informative training subsets. GradNorm prioritizes large-gradient
examples \citep{katharopoulos2018importance,wang2024greats}, while
GraNd and EL2N use early-training scores for dataset pruning
\citep{paul2021datadiet}.
GRAD-MATCH selects weighted coresets that approximate training or
validation gradients, aiming to preserve performance with fewer examples
\citep{killamsetty2021gradmatch}.
For LLMs, LESS selects instruction data offline \citep{xia2024less},
while GREATS performs online subset selection \citep{wang2024greats}.
Both use held-out target examples as the selection reference.
\method{} instead focuses on filtering synthetic supervision during
mixed real/synthetic training. It retains all real data and estimates
synthetic utility using a smoothed real-training gradient reference,
with history-calibrated batch utility determining when sample-level
filtering is needed.

\section{Method}
\label{sec:method}

\noindent\textbf{Problem setup.}
\label{sec:problem-setup}
We have access to a real training dataset $\mathcal D_R$ and a synthetic
dataset $\mathcal D_S$. At training step $t$, we draw a real mini-batch
$\mathcal R_t\subseteq\mathcal D_R$ and a synthetic mini-batch
$S_t=\{s_{t1},\ldots,s_{tm}\}\subseteq\mathcal D_S$.
Given the current model $f_{\theta_t}$, we select
$\widetilde S_t\subseteq S_t$ to train alongside $\mathcal R_t$.
Our goal is to reduce the expected loss on real test data drawn from the
target distribution $P_R$:
\begin{equation}
    \mathcal{L}_R(\theta)
    = \mathbb{E}_{r\sim P_R}\!\left[\ell(r;\theta)\right].
\end{equation}
Here, $\ell(r;\theta)$ is the per-example loss.
Selection uses the current model, available training data, and past training
statistics, without data quality labels or a separate held-out validation
set.

\begin{figure}[!t]
    \centering
    \includegraphics[width=\linewidth]{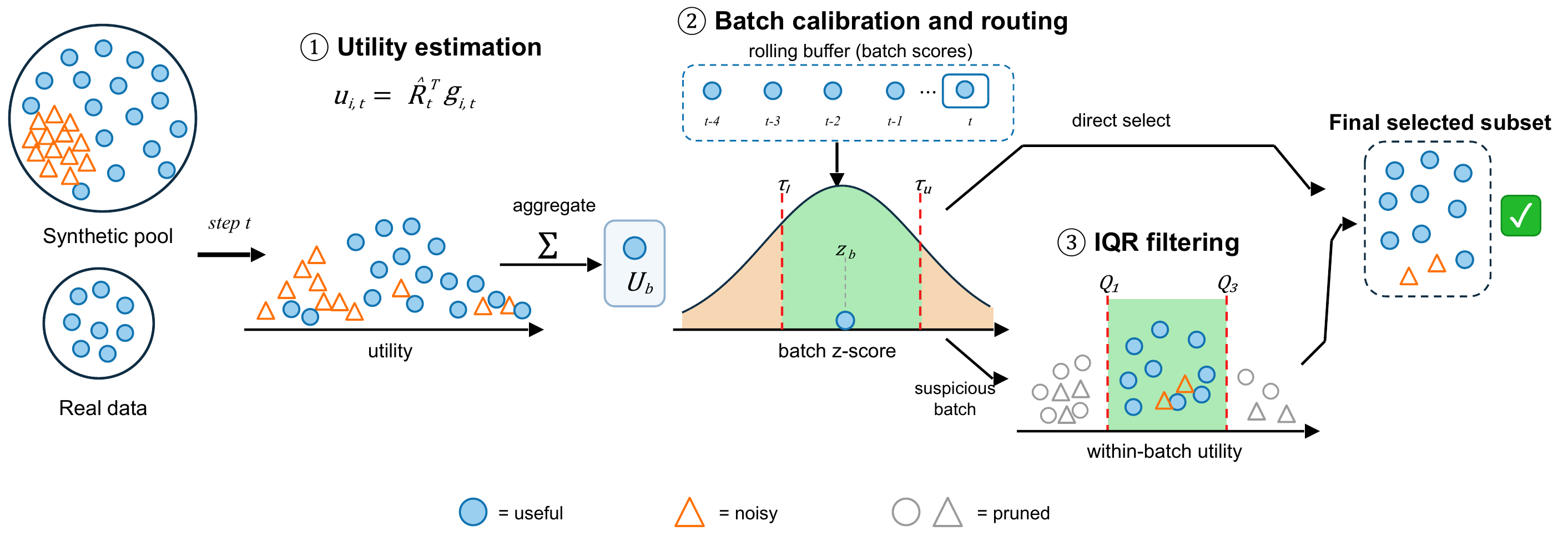}
    \caption{Overview of \method{}. The framework
    \circled{1} estimates each synthetic sample's utility through gradient alignment
    with a smoothed real-training reference,
    \circled{2} aggregates these scores and calibrates batch utility against recent
    training history to retain moderate-utility batches directly, and
    \circled{3} applies IQR-based sample filtering to out-of-band batches.}
    \label{fig:huf-overview}
\end{figure}

\noindent\textbf{Overview.}
Figure~\ref{fig:huf-overview} presents three components of \method{}.
\circled{1} We estimate each synthetic sample's utility through gradient
alignment with a smoothed real-training reference.
\circled{2} We aggregate these scores and calibrate batch utility against
recent training history, retaining batches within a moderate utility band.
\circled{3} For out-of-band batches, we apply sample-level filtering based
on the interquartile range (IQR).
We describe these three components in detail in
Sections~\ref{sec:sample-utility}--\ref{sec:iqr}.

\subsection{Estimating synthetic sample utility}
\label{sec:sample-utility}
Inspired by TracIn \citep{pruthi2020tracin}, we estimate a synthetic sample's
utility by asking how an update with that sample would change the current
model's real-data loss. Let $\theta_t^{(i)}$ denote the parameters after a
hypothetical update with $s_{ti}$, and write
$d_{i,t}=\theta_t^{(i)}-\theta_t$. For a locally smooth real-data loss, a
Taylor expansion around $\theta_t$ expresses the loss reduction as
\begin{equation}
    \begin{aligned}
        \Delta_{i,t}
        &=\mathcal L_R(\theta_t)-\mathcal L_R(\theta_t^{(i)})\\
        &=-\nabla_\theta\mathcal L_R(\theta_t)^\top d_{i,t}
          +O\!\left(\|d_{i,t}\|^2\right).
    \end{aligned}
    \label{eq:loss-change}
\end{equation}
For a hypothetical SGD step of size $\eta>0$,
$d_{i,t}=-\eta\nabla_\theta\ell(s_{ti};\theta_t)$. Substituting this update
into Equation~\ref{eq:loss-change} gives
\begin{equation}
    \Delta_{i,t}
    =\eta\nabla_\theta\mathcal L_R(\theta_t)^\top
       \nabla_\theta\ell(s_{ti};\theta_t)
     +O\!\left(\eta^2\|\nabla_\theta\ell(s_{ti};\theta_t)\|^2\right).
    \label{eq:first-order}
\end{equation}
Thus, positive alignment between the synthetic gradient and the target real gradient
predicts a first-order reduction in real-data loss.

\noindent\textbf{A reference from real training data.}
Measuring a real-test loss change for every candidate would require repeated
model updates and access to test labels. Instead, we estimate the target
gradient using the current real training mini-batch $\mathcal R_t$.
Let $\phi$ be the parameter block of $\theta$ used for scoring, with
$\phi=\theta$ corresponding to the full-model derivation above. Define
$g_t(x)=\nabla_\phi\ell(x;\theta_t)$ and the synthetic sample gradient
$g_{i,t}=g_t(s_{ti})$. The real reference is
\begin{equation}
    R_t=\frac{1}{|\mathcal R_t|}\sum_{r\in\mathcal R_t}g_t(r),
    \label{eq:real-reference}
\end{equation}
which estimates the target gradient in the same parameter block. This lets
us estimate local alignment through $R_t^\top g_{i,t}$ without separately
updating the model for each candidate.

\noindent\textbf{Smoothing the reference.}
A single real mini-batch can give a noisy reference. We smooth it across
steps using a bias-corrected EMA:
\begin{equation}
    M_t=\beta M_{t-1}+(1-\beta)R_t,\qquad
    \hat R_t=\frac{M_t}{1-\beta^t},\qquad
    M_0=0,\quad\beta=0.99,
    \label{eq:real-gradient-ema}
\end{equation}
where $t$ counts reference updates. Appendix~\ref{app:ema-tracking}
bounds the EMA tracking error in terms of mini-batch noise and
parameter drift. The sample score is the unnormalized inner product
\begin{equation}
    u_{i,t}=\hat R_t^\top g_{i,t}.
    \label{eq:sample-utility}
\end{equation}
In practice, we approximate full-model gradient alignment using a selected
parameter block: $\phi$ is the prediction head for image models and LLMs,
and the final layer of the shared backbone for the deep learning
recommendation model used in ads re-ranking.
The resulting
score is a proxy for the effect of a full-model update.
Computational overhead is analyzed in
Appendix~\ref{app:computational-overhead}.

\subsection{Batch-level calibration and routing}
\label{sec:batch-calibration}
Using the estimated sample utilities, we first assess the synthetic batch
as a whole. We calibrate its aggregate utility and use a bounded utility
band to decide whether to apply sample-level filtering, rather than
whether to discard the entire batch.

\noindent\textbf{Batch aggregation and calibration.}
Batch averaging provides a more stable scale for deciding when to apply
sample-level filtering. Define
\begin{equation}
    \bar g_{S,t}=\frac{1}{m}\sum_{i=1}^m g_{i,t},
    \qquad
    U_t=\frac{1}{m}\sum_{i=1}^m u_{i,t}
       =\hat R_t^\top\bar g_{S,t}.
    \label{eq:batch-utility}
\end{equation}
Conditioned on a fixed model and reference, suppose candidate utilities are
i.i.d.\ with mean $\mu_t$ and finite, nonzero variance $\sigma_t^2$.
The central limit theorem (CLT) gives
\begin{equation}
    \frac{\sqrt m\,(U_t-\mu_t)}{\sigma_t}
    \xrightarrow[m\to\infty]{d}\mathcal N(0,1),
    \qquad
    \operatorname{Var}(U_t)=\frac{\sigma_t^2}{m}.
    \label{eq:local-clt}
\end{equation}
Consequently, for sufficiently large batches, the batch mean admits a normal
approximation without assuming Gaussian sample-level scores.
When the model and reference change slowly over a short
training window, recent batch utilities provide a local distribution for
calibration. We keep up to $W$ preceding batch scores in a rolling buffer
$\mathcal B_t$. Its empirical mean and standard deviation define the
current batch's $z_t$-score:
\begin{equation}
    \hat\mu_t=\frac{1}{|\mathcal B_t|}\sum_{v\in\mathcal B_t}v,
    \qquad
    \hat\sigma_t^2=\frac{1}{|\mathcal B_t|}\sum_{v\in\mathcal B_t}(v-\hat\mu_t)^2,
    \qquad
    z_t=\frac{U_t-\hat\mu_t}{\hat\sigma_t+\epsilon},
    \label{eq:z-score}
\end{equation}
where $\epsilon=10^{-8}$. The buffer estimates the local distribution of
\emph{batch}-level utilities. During filtering, the current $U_t$ enters the buffer only
after its routing decision.

\noindent\textbf{Initialization.}
During the first approximately $5\%$ of training steps, only real data
update the model; we update the real-gradient EMA and score synthetic
batches to populate the buffer. Mixed real/synthetic training and
filtering begin after warmup, using the available buffer entries.

\noindent\textbf{Bounded utility and batch routing.}
\label{sec:routing}
Batch utility $U_t$ provides a first-order estimate of how a synthetic
update affects the real-data objective. Reference estimation error and
higher-order effects can change the realized loss reduction, so the score
indicates local update alignment rather than guaranteeing training benefit.
Appendix~\ref{app:utility-approximation} gives the decomposition and error
bound. Our empirical study in Section~\ref{sec:utility-horizon} also shows that 
the high-utility band yields a higher early performance peak but fails to sustain this advantage, leaving lower mean performance
than the moderate band. The moderate band provides a better balance between short-term performance and long-term stability. 
Thus, we use a bounded interval of the calibrated score $z_t$ as a
trust region for direct acceptance of moderate-utility batches. The lower bound flags relatively weak aggregate utility. Finally, a batch
is retained in full when
\begin{equation}
    \tau_l\le z_t\le\tau_u,\qquad \tau_l<\tau_u.
    \label{eq:trust-region}
\end{equation}
The band is moderate relative to recent batches. Batches outside either
bound are routed to sample-level filtering.

\subsection{Sample-level IQR filtering}
\label{sec:iqr}
For batches routed outside the utility band, we refine the decision at the
sample level: an out-of-band average alone does not identify which samples
to remove. We therefore inspect their within-batch utilities. Unlike batch
averages, individual scores do not inherit a Gaussian approximation from
the CLT, so we use empirical quartiles to set robust filtering thresholds.
We use IQR-scaled thresholds based on Tukey's fences
\citep{tukey1977exploratory}, with separate lower and upper multipliers.
Let $Q_{1,t}$ and $Q_{3,t}$ be the $25$th and $75$th percentiles of
$\{u_{i,t}\}_{i=1}^m$, and let $I_t=Q_{3,t}-Q_{1,t}$ be their IQR. Define
\begin{equation}
    a_t=Q_{1,t}-\lambda_l I_t,\qquad
    b_t=Q_{3,t}+\lambda_u I_t,
    \qquad \lambda_u\ge0,\quad a_t\le b_t.
    \label{eq:iqr}
\end{equation}
A negative $\lambda_l$ tightens the lower cutoff to
$Q_{1,t}+|\lambda_l|I_t$. These thresholds scale with the central spread
rather than the magnitudes of extreme scores. The complete selection rule is
\begin{equation}
    \widetilde S_t=
    \begin{cases}
        S_t, & \tau_l\le z_t\le\tau_u,\\
        \{s_{ti}\in S_t:a_t\le u_{i,t}\le b_t\},
            & \text{otherwise}.
    \end{cases}
    \label{eq:routing}
\end{equation}
The retained samples train alongside $\mathcal R_t$.

\section{Experiments}
\label{sec:experiments}

We evaluate \method{} on image classification, LLM fine-tuning for
text-to-SQL, and industrial ads re-ranking. We compare real-task
performance with existing selection methods and examine the two-stage
design through ablations and a batch-utility analysis.

The image and LLM experiments use public datasets and share the same
batch-gate bounds
$(\tau_l,\tau_u)=(-0.025,1)$ and IQR multipliers
$(\lambda_l,\lambda_u)=(0,1.5)$, without domain- and model-specific tuning of these
parameters. The synthetic-data drop ratio is the percentage of synthetic
data filtered out from training. Real-only and unfiltered all synthetic data training correspond to drop ratios of
$100\%$ and $0\%$, respectively.

\subsection{Image classification}
\label{sec:image-experiments}

\noindent\textbf{Data and model.}
We first evaluate \method{} on CIFAR-100 image classification
\citep{krizhevsky2009learning} using ResNet-18 \citep{he2016resnet}.
CIFAR-100 contains 100 classes with 500 images per class, and we evaluate accuracy on 10,000 real test images. We use FLUX.1~[schnell] \citep{blackforestlabs2024flux1schnell}, SANA \citep{xie2025sana}, and Stable Diffusion v1.4 (SD1.4) \citep{rombach2022ldm} to generate synthetic images for training. For FLUX.1~[schnell] and SANA, we generate 2,000 synthetic images per class. For SD1.4, we directly use the dataset released by \citet{shipard2023diversity} which contains 1,800 images per class. We combine each synthetic pool separately with the real training data to evaluate the effectiveness of \method{} across different synthetic data sources. Training and synthetic-data
generation details are provided in
Appendix~\ref{app:image-implementation}.

\noindent\textbf{Baselines.}
We compare \method{} with \emph{Random}, which uniformly downsamples the
synthetic pool, and two groups of selection methods. For dynamic data
selection, \emph{InfoBatch} randomly prunes a portion of low-loss examples
and rescales retained gradients to correct sampling bias
\citep{qin2024infobatch}. \emph{OrderDP} retains examples with the highest
loss-based scores from randomly sampled candidate subsets
\citep{jin2026orderdp}.
For synthetic-data selection, we use two feature-based methods as baselines: \emph{DS3} samples from synthetic feature clusters nearest to real examples, while \emph{CovMatch} greedily selects examples to match the subset's feature covariance to
that of real data
\citep{hulkund2025datas3,rezaei2026highdimensional}.
We also include \emph{Real only} and \emph{Real + All Synthetic} as
references. All methods retain the real training data, and selection
baselines match \method{}'s synthetic-data drop ratio for each generator.

\noindent\textbf{Selection results across generators.}
As shown in Table~\ref{tab:cross-generator}, \method{} outperforms both
dynamic data selection baselines and methods tailored to synthetic-data
selection across all three generators at matched drop ratios. The dynamic
baselines \emph{InfoBatch} and \emph{OrderDP} target training-data
efficiency, whereas \emph{DS3} and \emph{CovMatch}
emphasize feature-space coverage or matching real-data statistics.
\method{} instead filters synthetic data using a utility signal tied to
the real-data training objective to retain more beneficial supervision.
Compared with full synthetic-pool training, absolute accuracy improves by
$0.58\%$, $0.65\%$, and $1.14\%$ on FLUX, SANA, and SD1.4, respectively.
\begin{table}[H]
    \caption{CIFAR-100 classification with synthetic training data. $\Delta$ is the accuracy change relative to Real + All Synthetic for the same generator.}
    \label{tab:cross-generator}
    \centering
    \small
    \setlength{\tabcolsep}{3pt}
    \begin{tabular*}{\linewidth}{@{\extracolsep{\fill}}lccccccccc@{}}
        \toprule
        & \multicolumn{3}{c}{FLUX}
        & \multicolumn{3}{c}{SANA}
        & \multicolumn{3}{c}{SD1.4} \\
        \cmidrule(lr){2-4}\cmidrule(lr){5-7}\cmidrule(lr){8-10}
        Method & \textit{Drop} & Acc. $\uparrow$ & $\Delta\uparrow$
        & \textit{Drop} & Acc. $\uparrow$ & $\Delta\uparrow$
        & \textit{Drop} & Acc. $\uparrow$ & $\Delta\uparrow$ \\
        \midrule
        Real only & \textit{100} & 76.04 & -- & \textit{100} & 76.04 & -- & \textit{100} & 76.04 & -- \\
        Real + All Synthetic & \textit{0} & 80.31 & $0.00$ & \textit{0} & 79.32 & $0.00$ & \textit{0} & 79.46 & $0.00$ \\
        \midrule
        Random & \textit{25.76} & 80.33 & $+0.02$ & \textit{26.01} & 79.61 & $+0.29$ & \textit{24.58} & 79.34 & $-0.12$ \\
        DS3 & \textit{25.76} & 80.01 & $-0.30$ & \textit{26.01} & 79.77 & $+0.45$ & \textit{24.58} & 79.44 & $-0.02$ \\
        Covariance Matching & \textit{25.76} & 79.92 & $-0.39$ & \textit{26.01} & 79.21 & $-0.11$ & \textit{24.58} & 79.21 & $-0.25$ \\
        InfoBatch & \textit{25.76} & 79.84 & $-0.47$ & \textit{26.01} & 78.94 & $-0.38$ & \textit{24.58} & 79.24 & $-0.22$ \\
        OrderDP & \textit{25.76} & 80.36 & $+0.05$ & \textit{26.01} & 79.71 & $+0.39$ & \textit{24.58} & 79.66 & $+0.20$ \\
        \midrule
        \method{} (ours) & \textit{25.76} & \textbf{80.89} & $\boldsymbol{+0.58}$ & \textit{26.01} & \textbf{79.97} & $\boldsymbol{+0.65}$ & \textit{24.58} & \textbf{80.60} & $\boldsymbol{+1.14}$ \\
        \bottomrule
    \end{tabular*}
\end{table}

\begingroup
\setlength{\intextsep}{6pt}
\setlength{\columnsep}{10pt}
\Needspace{13\baselineskip}
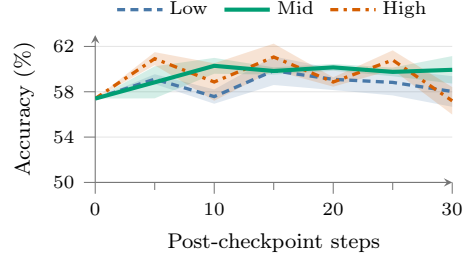
\begin{wrapfigure}[12]{r}{0.38\textwidth}
    \centering
    \begingroup
\definecolor{huflow}{HTML}{4878A8}
\definecolor{hufmid}{HTML}{009E73}
\definecolor{hufhigh}{HTML}{D55E00}

\newcommand{\hufutilitycolumns}[1]{\pgfplotstablecreatecol[
        create col/expr={
            (\thisrow{#1_r0}+\thisrow{#1_r1}+\thisrow{#1_r2})/3
        }
    ]{#1_mean}\hufutilitydata
    \pgfplotstablecreatecol[
        create col/expr={
            sqrt((
                (\thisrow{#1_r0}-\thisrow{#1_mean})^2
               +(\thisrow{#1_r1}-\thisrow{#1_mean})^2
               +(\thisrow{#1_r2}-\thisrow{#1_mean})^2
            )/6)
        }
    ]{#1_sem}\hufutilitydata
}

\IfFileExists{figures/utility_accuracy_steps0_30.csv}{\def\hufutilitydataready{1}
    \pgfplotstableread[col sep=comma]
        {figures/utility_accuracy_steps0_30.csv}\hufutilitydata
    \hufutilitycolumns{low}
    \hufutilitycolumns{mid}
    \hufutilitycolumns{high}
}{\def\hufutilitydataready{0}
}

\newcommand{\hufutilityshade}[2]{\addplot[draw=none, name path=huf-#1-upper, forget plot]
        table[x=step, y expr=\thisrow{#1_mean}+\thisrow{#1_sem}]
        {\hufutilitydata};
    \addplot[draw=none, name path=huf-#1-lower, forget plot]
        table[x=step, y expr=\thisrow{#1_mean}-\thisrow{#1_sem}]
        {\hufutilitydata};
    \addplot[draw=none, fill=#2, fill opacity=0.17, forget plot]
        fill between[of=huf-#1-upper and huf-#1-lower];
}

\begin{tikzpicture}
    \begin{axis}[
        width=\linewidth,
        height=0.56\linewidth,
        xmin=0, xmax=30,
        ymin=50, ymax=63,
        xtick={0,10,20,30},
        minor xtick={5,15,25},
        ytick={50,54,58,62},
        xlabel={Post-checkpoint steps},
        ylabel={Accuracy (\%)},
        label style={font=\footnotesize},
        tick label style={font=\scriptsize},
        axis lines=left,
        axis line style={black!55},
        tick align=outside,
        tick style={black!55},
        ymajorgrids=true,
        grid style={black!10, thin},
        legend columns=3,
        legend style={
            at={(0.5,1.04)}, anchor=south,
            draw=none, fill=none,
            font=\scriptsize,
            /tikz/every even column/.append style={column sep=0.35em}
        },
        clip=true
    ]
        \addlegendimage{huflow, line width=1.25pt, densely dashed}
        \addlegendentry{Low}
        \addlegendimage{hufmid, line width=1.55pt, solid}
        \addlegendentry{Mid}
        \addlegendimage{hufhigh, line width=1.25pt, dashdotted}
        \addlegendentry{High}

        \ifnum\hufutilitydataready=1\relax
            \hufutilityshade{low}{huflow}
            \hufutilityshade{mid}{hufmid}
            \hufutilityshade{high}{hufhigh}
            \addplot[huflow, line width=1.25pt, densely dashed, forget plot]
                table[x=step, y=low_mean] {\hufutilitydata};
            \addplot[hufhigh, line width=1.25pt, dashdotted, forget plot]
                table[x=step, y=high_mean] {\hufutilitydata};
            \addplot[hufmid, line width=1.55pt, solid, forget plot]
                table[x=step, y=mid_mean] {\hufutilitydata};
        \else
            \node[
                align=center, text=black!65, fill=white,
                inner sep=7pt, font=\small
            ] at (rel axis cs:0.5,0.52) {
                \textbf{TBD: per-repeat accuracy data}\\[3pt]
                No experimental curves plotted.
            };
        \fi
    \end{axis}
\end{tikzpicture}
\endgroup
    \caption{Test accuracy over following 30 steps.}
    \label{fig:utility-early-horizon}
\end{wrapfigure}
\noindent\textbf{Effect of batch utility.}
\label{sec:utility-horizon}
We examine whether higher batch utility leads to steadier training and
better classification accuracy. We first train the model for 20 epochs. Holding the
checkpoint and all other state variables fixed, we sample synthetic batches from Low
($z<-0.025$), Mid ($-0.025\le z\le1$), and High ($z>1$) utility bands.
Each branch starts from the same checkpoint and optimizer state, with 200
real and 800 synthetic images per step. We compare the first 30 continuation steps
over 3 repeats. Figure~\ref{fig:utility-early-horizon} shows that
High reaches a higher early accuracy peak but exhibits larger fluctuations
in its mean accuracy curve. Mid is steadier and finishes this window with
higher mean accuracy ($59.94\%$ versus $57.20\%$ for High at step 30).
Thus, a larger current utility does not necessarily yield steadier or
better training results.
\par
\endgroup

\subsection{LLM fine-tuning for text-to-SQL}
\label{sec:llm-experiments}

\noindent\textbf{Data and evaluation.}
We next apply \method{} to LLM fine-tuning for the text-to-SQL task.
Our real training data consist of 8,659 human-annotated question--SQL
pairs from Spider~1.0 \citep{yu2018spider}. Our synthetic data come from
SynSQL, which contains automatically generated question--SQL pairs over
synthetic databases \citep{li2025omnisql}. We sample a subset of 34,636
examples from SynSQL as our synthetic training pool, 4 times the size
of the real training set, matching the $4{:}1$ ratio in our FLUX and SANA
image settings. We evaluate on Spider~1.0 dev and report execution
accuracy (EX), the fraction of predicted SQL queries whose execution
results match those of the gold queries, alongside the synthetic-data
drop ratio.

\noindent\textbf{Model.}
We fully fine-tune Qwen2.5-Coder-0.5B-Instruct
\citep{hui2024qwen25coder} for one epoch with a learning rate of
$5\times10^{-5}$ and a global batch size of 64. Utility scores use
gradient alignment at the LM output head against the real-gradient EMA
reference. Training and selection implementation details are provided in
Appendix~\ref{app:llm-implementation}.

\noindent\textbf{Baselines.}
We compare \method{} with \emph{Random} downsampling, \emph{GradNorm},
which prioritizes examples with the largest per-example gradient norms,
motivated by gradient-norm-based importance sampling
\citep{katharopoulos2018importance}, and \emph{GREATS}
\citep{wang2024greats}, which uses a Taylor approximation to select
examples that reduce a reference loss. For a fair comparison, our
GREATS implementation uses real-training gradients as the reference
instead of held-out validation gradients. We also include
\emph{Real only} and \emph{Real + All Synthetic} to measure performance
without augmentation and with the unfiltered SynSQL pool, respectively.
Random, GradNorm, and GREATS use the same $20.4\%$ drop ratio as
\method{} to match the synthetic-data budget.

\noindent\textbf{Synthetic-data selection results.}
Table~\ref{tab:llm-results} reports overall EX and its breakdown by
difficulty. \method{} reaches $65.7\%$ EX while dropping $20.4\%$ of the
synthetic data, outperforming the other data selection methods at the
same drop ratio. It also improves EX by $0.9\%$ over
training with the full synthetic pool.
\begin{table}[H]
    \caption{Spider~1.0 fine-tuning with synthetic data pool, evaluated on the dev split. EX is execution accuracy (\%), reported overall and by difficulty. $\Delta$EX is the overall EX change relative to Real + All Synthetic. All methods retain the real training data; selection methods use a matched drop ratio.}
    \label{tab:llm-results}
    \centering
    \small
    \setlength{\tabcolsep}{3pt}
    \begin{tabular*}{\linewidth}{@{\extracolsep{\fill}}lccccccc@{}}
        \toprule
        & & & & \multicolumn{4}{c}{EX by difficulty (\%) $\uparrow$} \\
        \cmidrule(lr){5-8}
        Method & \textit{Drop} & EX $\uparrow$ & $\Delta$EX $\uparrow$ & Easy & Medium & Hard & Extra Hard \\
        \midrule
        Real only & \textit{100} & 57.1 & -- & 79.4 & 59.6 & 43.7 & 30.7 \\
        Real + All Synthetic & \textit{0} & 64.8 & $0.0$ & 83.5 & 67.0 & 56.9 & 39.2 \\
        \midrule
        Random & \textit{20.4} & 64.1 & $-0.7$ & 83.9 & 67.5 & 51.7 & 38.6 \\
        GradNorm & \textit{20.4} & 64.2 & $-0.6$ & 82.3 & 67.7 & 55.7 & 36.7 \\
        GREATS & \textit{20.4} & 65.3 & $+0.5$ & 81.9 & 68.4 & 57.5 & 40.4 \\
        \midrule
        \method{} (ours) & \textit{20.4} & \textbf{65.7} & $\mathbf{+0.9}$ & 83.5 & 68.6 & 54.6 & 42.8 \\
        \bottomrule
    \end{tabular*}
\end{table}

\begingroup
\setlength{\intextsep}{6pt}
\setlength{\columnsep}{10pt}
\begin{wrapfigure}[11]{r}{0.43\textwidth}
    \centering
    \begingroup
\definecolor{hufsweep}{HTML}{4878A8}
\colorlet{hufgood}{green!45!black}

\begin{tikzpicture}
    \begin{axis}[
        width=\linewidth,
        height=3.0cm,
        ymin=0, ymax=11.4,
        ytick={0,2,4,6,8,10},
        ylabel={EX change (pp) $\uparrow$},
        symbolic x coords={pad0,allsyn,batch,sample,huf,pad1},
        xmin=pad0, xmax=pad1,
        xtick={allsyn,batch,sample,huf},
        xticklabels={{All syn.\\0\%},{Batch\\80.0\%},
                     {Sample\\32.2\%},{\textbf{\method{}}\\20.4\%}},
        xticklabel style={align=center, font=\scriptsize},
        enlarge x limits=0,
        label style={font=\scriptsize},
        tick label style={font=\scriptsize},
        axis lines=left,
        axis line style={black!55},
        tick align=outside,
        tick style={black!55},
        ymajorgrids=true,
        grid style={black!10, thin},
        clip=false,
        nodes near coords,
        every node near coord/.append style={
            font=\scriptsize,
            fill=white, inner sep=0.7pt,
            /pgf/number format/fixed,
            /pgf/number format/fixed zerofill,
            /pgf/number format/print sign,
            /pgf/number format/precision=1
        },
    ]
        \draw[black!55, densely dashed, line width=0.7pt]
            ({rel axis cs:0,0} |- {axis cs:allsyn,7.7})
            -- ({rel axis cs:1,0} |- {axis cs:allsyn,7.7});

        \addplot[
            ybar, bar width=14pt, bar shift=0pt,
            fill=black!12, draw=black!55, line width=0.8pt,
            every node near coord/.append style={text=black!60}
        ] coordinates {(allsyn,7.7)};

        \addplot[
            ybar, bar width=14pt, bar shift=0pt,
            fill=hufsweep!22, draw=hufsweep, line width=0.8pt,
            every node near coord/.append style={text=hufsweep}
        ] coordinates {(batch,3.6)};

        \addplot[
            ybar, bar width=14pt, bar shift=0pt,
            fill=hufsweep!22, draw=hufsweep, line width=0.8pt,
            every node near coord/.append style={text=hufsweep}
        ] coordinates {(sample,6.0)};

        \addplot[
            ybar, bar width=14pt, bar shift=0pt,
            fill=hufgood!25, draw=hufgood, line width=1.0pt,
            every node near coord/.append style={text=hufgood}
        ] coordinates {(huf,8.6)};

        \node[text=hufgood, font=\small\bfseries, anchor=south]
            at (axis cs:huf,10.5) {$\checkmark$};
    \end{axis}
\end{tikzpicture}
\endgroup
    \caption{Two-stage ablation on Spider~1.0 dev, measured against the
    real-only model.}
    \label{fig:llm-stage-combination}
\end{wrapfigure}
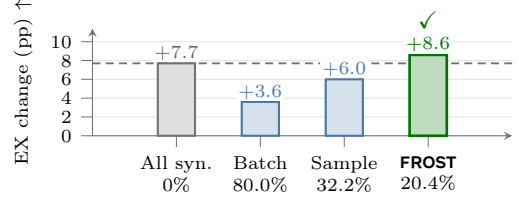
\noindent\textbf{Batch-level and sample-level ablations.}
We examine how the two stages affect synthetic-data use and downstream EX
(Figure~\ref{fig:llm-stage-combination}). \emph{Batch gate only} retains
batches within the utility band and drops out-of-band batches without
sample-level filtering. \emph{Sample IQR only} applies the IQR rule to
every synthetic batch without batch-level routing. Batch gate only drops
$80\%$ of the synthetic data and underperforms full \method{}, suggesting
that out-of-band batches still contain useful samples that sample-level
IQR filtering can retain. Sample IQR only also underperforms full
\method{}: filtering every batch can remove useful supervision from
batches that do not need cleanup.
\par
\endgroup

\subsection{Ads re-ranking on real-world data}
\label{sec:ads-experiments}

\noindent\textbf{Data and task.}
Finally, we evaluate \method{} on a production ads re-ranking task, focusing
on conversion prediction. Each real training example is an ad impression
paired with a binary label indicating whether an advertiser-defined
conversion event, such as a purchase, an app install, or a sign-up, is
\emph{attributed} to that impression. Attribution links an observed
conversion to the impression credited with it, providing a positive
training label. We train on a 14-day data window ending on day $T$,
containing billions of real impressions, and evaluate offline on a
24-hour window on day $T+2$.

\noindent\textbf{Synthetic-data generation.}
To expand and enrich the training data, we generate synthetic examples from
\emph{unattributed} conversions:
observed conversions that cannot be tied back to a specific impression because of
privacy-related attribution gaps. We sample likely impressions based on
user--ad interaction frequencies and pair them with these conversions to
create synthetic examples. The resulting synthetic pool is about $\frac{1}{3}$ the size of the real training set. The conversion events are
observed, but their links to impressions are inferred and lack
ground-truth verification. Ads re-ranking
models are sensitive to this label noise because incorrect
impression--conversion pairs can distort the learned conversion
probabilities. In our experiments, training with all synthetic examples
regresses relative to real-only training (Table~\ref{tab:ads-results}).
We therefore apply \method{} during mixed real/synthetic training to
select synthetic examples that provide useful supervision for the
real-data task.

\noindent\textbf{Model.}
We use a deep learning recommendation model to predict conversion
probabilities.

\noindent\textbf{Evaluation metric: Normalized Entropy.}
We use Normalized Entropy (NE) to evaluate the model's binary predictions \citep{he2014predictingclicks}.
NE normalizes the average binary cross-entropy by the label entropy
computed from the observed conversion rate in the evaluation set:
\begin{equation}
    \mathrm{NE}
    =\frac{-\frac{1}{N}\sum_{i=1}^{N}
        \left[y_i\log p_i+(1-y_i)\log(1-p_i)\right]}
        {-\hat p\log\hat p-(1-\hat p)\log(1-\hat p)}.
    \label{eq:normalized-entropy}
\end{equation}
Here, $N$ is the number of real evaluation examples, $y_i\in\{0,1\}$ is
the observed label, $p_i$ is the model's predicted probability, and
$\hat p=N^{-1}\sum_{i=1}^{N}y_i$ is the observed conversion rate.
Lower NE indicates better probabilistic predictions, and even small
reductions can be meaningful at industrial scale
\citep{he2014predictingclicks}.
In well-optimized commercial recommendation systems, relative NE reduction
of only $0.02\%$ has been considered significant in prior industrial
studies \citep{li2022frequencyaware,lai2023adaembed}.

We report relative NE change as
$\Delta\mathrm{NE}\,(\%)=100(\mathrm{NE}_{\mathrm{model}}/\mathrm{NE}_{\mathrm{real}}-1)$,
where $\mathrm{NE}_{\mathrm{model}}$ and $\mathrm{NE}_{\mathrm{real}}$
are the NE of the evaluated model and the fixed real-only baseline,
respectively. All models are evaluated on the same real-data split using
the same $\hat p$. The real-only baseline is $0\%$; negative values indicate
improvement and positive values indicate regression.

\noindent\textbf{Effect of synthetic-data selection.}
We include \emph{Random} downsampling at $30\%$ and $70\%$ drop
ratios as a general-purpose selection baseline. Due to constraints
of the commercial setup, more comprehensive baseline comparisons are
conducted on the public image-classification and text-to-SQL tasks.
Table~\ref{tab:ads-results} shows that adding all
synthetic data \textcolor{red!80!black}{\textbf{increases}} NE by
\textcolor{red!80!black}{$0.211\%$} compared to real-only training.
Although synthetic data expands the training set, noise in the inferred
labels can offset this benefit and hurt performance. Applying \textbf{\method{}}
\textcolor{green!45!black}{\textbf{lowers}} NE by
\textcolor{green!45!black}{$0.096\%$} while
dropping only $36\%$ of the synthetic data. \method{} therefore reverses the
effect of the synthetic pool, converting a
\textcolor{red!80!black}{$0.211\%$} regression into a
\textcolor{green!45!black}{$0.096\%$} improvement.
This is direct evidence that \method{} filters out the noisy and harmful synthetic
data and retains useful examples that contribute genuine signal to the model,
with an NE gain far beyond the industrial significance threshold.
\begingroup
\setlength{\intextsep}{6pt}
\begin{table}[!htbp]
    \caption{Ads re-ranking results relative to real-only training. \textit{Drop} is the synthetic-data drop ratio; all real data are retained. \textcolor{red!80!black}{Red $\uparrow$} denotes higher NE (worse); \textcolor{green!45!black}{green $\downarrow$} denotes lower NE (better).}
    \label{tab:ads-results}
    \centering
    \small
    \begin{tabular}{lcc}
        \toprule
        Training data / selection & \textit{Drop} (\%) & Relative NE change (\%) $\downarrow$ \\
        \midrule
        Real data only (reference) & \textit{100} & -- \\
        $+$ All synthetic data & \textit{0} & \textcolor{red!80!black}{$+0.211\;\uparrow$} \\
        $+$ Random & \textit{30} & \textcolor{red!80!black}{$+0.118\;\uparrow$} \\
        $+$ Random & \textit{70} & \textcolor{red!80!black}{$+0.023\;\uparrow$} \\
        \midrule
        \method{} & \textit{36} & \textcolor{green!45!black}{$-0.096\;\downarrow$} \\
        \bottomrule
    \end{tabular}
\end{table}

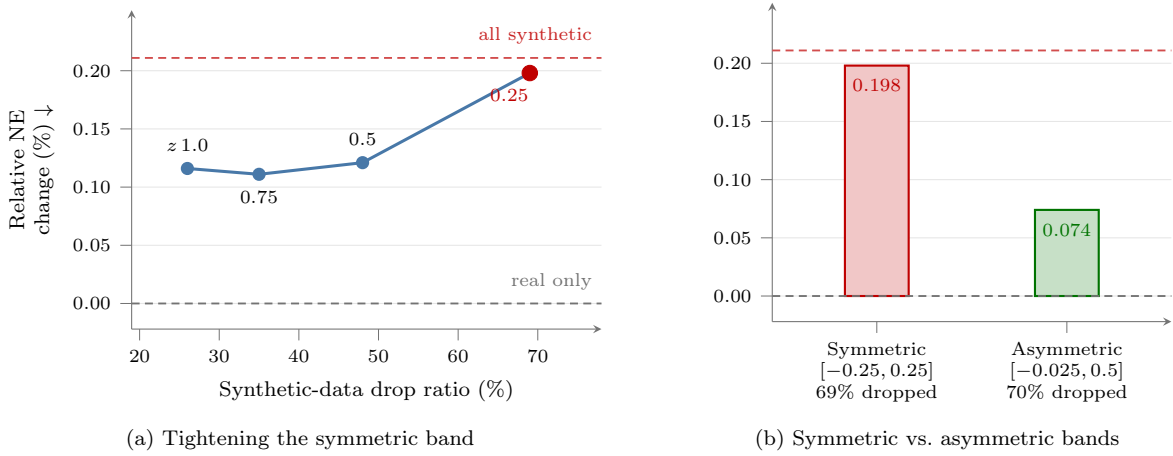
\begin{figure}[H]
    \centering
    \begingroup
\definecolor{hufband}{HTML}{4878A8}
\colorlet{hufbad}{red!75!black}
\colorlet{hufgood}{green!45!black}

\pgfplotsset{
    huf gate axis/.style={
        height=5.8cm,
        ymin=-0.022, ymax=0.252,
        ytick={0,0.05,0.10,0.15,0.20},
        yticklabel style={/pgf/number format/fixed,
                          /pgf/number format/fixed zerofill,
                          /pgf/number format/precision=2},
        label style={font=\footnotesize},
        tick label style={font=\scriptsize},
        axis lines=left,
        axis line style={black!55},
        tick align=outside,
        tick style={black!55},
        ymajorgrids=true,
        grid style={black!10, thin},
        clip=false
    }
}

\begin{minipage}[t]{0.5\linewidth}
    \centering
    \begin{tikzpicture}
        \begin{axis}[
            huf gate axis,
            width=0.94\linewidth,
            xmin=19, xmax=78,
            xtick={20,30,40,50,60,70},
            xlabel={Synthetic-data drop ratio (\%)},
            ylabel={Relative NE\\change (\%) $\downarrow$},
            ylabel style={align=center},
        ]
            \addplot[hufbad!70, densely dashed, line width=0.7pt, forget plot]
                coordinates {(19,0.211) (78,0.211)};
            \node[font=\scriptsize, text=hufbad!85, anchor=south east]
                at (axis cs:78,0.215) {all synthetic};

            \addplot[black!55, densely dashed, line width=0.7pt, forget plot]
                coordinates {(19,0) (78,0)};
            \node[font=\scriptsize, text=black!55, anchor=south east]
                at (axis cs:78,0.004) {real only};

            \addplot[
                hufband, line width=1.15pt,
                mark=*, mark size=1.9pt,
                mark options={fill=hufband, draw=hufband}
            ] coordinates {(26,0.116) (35,0.111) (48,0.121) (69,0.198)};

            \addplot[
                only marks, mark=*, mark size=2.9pt,
                mark options={fill=hufbad, draw=hufbad}
            ] coordinates {(69,0.198)};

            \node[font=\scriptsize, anchor=south] at (axis cs:26,0.123)
                {$z\,1.0$};
            \node[font=\scriptsize, anchor=north] at (axis cs:35,0.105)
                {$0.75$};
            \node[font=\scriptsize, anchor=south] at (axis cs:48,0.128)
                {$0.5$};
            \node[font=\scriptsize, text=hufbad, anchor=north east]
                at (axis cs:70,0.192) {$0.25$};
        \end{axis}
    \end{tikzpicture}\\[1pt]
    {\footnotesize (a) Tightening the symmetric band}
\end{minipage}
\hfill
\begin{minipage}[t]{0.47\linewidth}
    \centering
    \begin{tikzpicture}
        \begin{axis}[
            huf gate axis,
            width=0.88\linewidth,
            symbolic x coords={sym,asym},
            xtick={sym,asym},
            xticklabels={{Symmetric\\$[-0.25,0.25]$\\69\% dropped},
                         {Asymmetric\\$[-0.025,0.5]$\\70\% dropped}},
            xticklabel style={align=center, font=\scriptsize},
            enlarge x limits=0.55,
        ]
            \addplot[
                ybar, bar width=24pt, bar shift=0pt,
                fill=hufbad!22, draw=hufbad, line width=0.8pt,
                nodes near coords,
                every node near coord/.append style={
                    font=\scriptsize, text=hufbad, anchor=north,
                    yshift=-1.5pt,
                    /pgf/number format/fixed,
                    /pgf/number format/precision=3
                }
            ] coordinates {(sym,0.198)};

            \addplot[
                ybar, bar width=24pt, bar shift=0pt,
                fill=hufgood!22, draw=hufgood, line width=0.8pt,
                nodes near coords,
                every node near coord/.append style={
                    font=\scriptsize, text=hufgood, anchor=north,
                    yshift=-1.5pt,
                    /pgf/number format/fixed,
                    /pgf/number format/precision=3
                }
            ] coordinates {(asym,0.074)};

            \draw[hufbad!70, densely dashed, line width=0.7pt]
                ({rel axis cs:0,0} |- {axis cs:sym,0.211})
                -- ({rel axis cs:1,0} |- {axis cs:sym,0.211});
            \draw[black!55, densely dashed, line width=0.7pt]
                ({rel axis cs:0,0} |- {axis cs:sym,0})
                -- ({rel axis cs:1,0} |- {axis cs:sym,0});
        \end{axis}
    \end{tikzpicture}\\[1pt]
    {\footnotesize (b) Symmetric vs.\ asymmetric bands}
\end{minipage}
\endgroup
    \caption{Sensitivity to the batch utility band on the ads model. Relative NE change (\%) uses the real-only reference; lower is better. Dashed lines mark real-only training ($0\%$) and all synthetic data ($+0.211\%$). (a) Narrowing the symmetric $z$-band does not consistently improve NE. (b) The asymmetric band achieves lower NE at a similar drop ratio.}
    \label{fig:ads-batch-gate}
\end{figure}
\endgroup

\noindent\textbf{Sensitivity to the batch utility band.}
We examine how the width and asymmetry of the $z$-band affect performance
in the batch-only variant. We progressively narrow a symmetric band
from $[-1,1]$ to $[-0.25,0.25]$.
Figure~\ref{fig:ads-batch-gate}(a) shows that the drop ratio rises from
$26\%$ to $69\%$, but dropping more data does not consistently improve
performance. This pattern suggests that aggressive whole-batch rejection
removes useful supervision along with noise: an out-of-band mean does
not imply that every sample in the batch is unhelpful.
We then compare the symmetric band $[-0.25,0.25]$ with the
asymmetric band $[-0.025,0.5]$. At similar drop ratios ($69\%$ and
$70\%$), the asymmetric band yields a lower relative NE change
($+0.074\%$ versus $+0.198\%$), as shown in
Figure~\ref{fig:ads-batch-gate}(b). Raising the lower bound excludes more
relatively low-utility batches, while relaxing the upper bound preserves
moderately high-utility batches that the symmetric gate would discard.
The upper bound still excludes extreme positive scores from direct
acceptance. This comparison supports favoring moderately positive
standardized utility rather than narrowing the band symmetrically around
zero. We use $[-0.025,0.5]$ for ads and $[-0.025,1]$ for both public domains.

\noindent\textbf{Batch-level and sample-level ablations.}
Figure~\ref{fig:ads-stage-combination} compares each stage in isolation
against their combination. \emph{Batch gate only} applies the asymmetric
gate and discards every out-of-band batch whole, leaving
\textcolor{red!80!black}{$+0.074\%$} while
dropping $70\%$ of the synthetic data. Given the noisy synthetic ads
supervision, we use a negative lower IQR multiplier ($\lambda_l<0$) to
tighten the acceptance interval. \emph{Sample IQR only} skips routing
and applies this IQR rule to every synthetic batch, reaching
\textcolor{red!80!black}{$+0.036\%$} at a
$54\%$ drop.

\begingroup
\Needspace{14\baselineskip}
\noindent
\begin{minipage}[t]{0.62\textwidth}
\vspace{0pt}
Full \method{} instead
retains moderate-utility batches in full and applies sample-level filtering
only to batches flagged by routing. This combination reaches
\textcolor{green!45!black}{$-0.096\%$}
while dropping only $36\%$ of the synthetic data, improving on real-only
training. These results confirm the value of batch-level routing:
applying IQR filtering to every batch can remove helpful samples and
hurt performance. Retaining moderate-utility batches entirely while applying
sample-level filtering to out-of-band batches yields better performance
than either stage alone.
\end{minipage}\hfill
\begin{minipage}[t]{0.36\textwidth}
    \vspace{0pt}
    \centering
    \ifdefined\captionsetup\captionsetup{hypcap=false}\fi
    \begingroup
\definecolor{hufsweep}{HTML}{4878A8}
\colorlet{hufbad}{red!75!black}
\colorlet{hufgood}{green!45!black}

\begin{tikzpicture}
    \begin{axis}[
        width=0.94\linewidth,
        height=3.6cm,
        ymin=-0.18, ymax=0.255,
        ytick={-0.10,0,0.10,0.20},
        yticklabel style={/pgf/number format/fixed,
                          /pgf/number format/fixed zerofill,
                          /pgf/number format/precision=2},
        ylabel={Relative NE\\change (\%) $\downarrow$},
        ylabel style={align=center},
        symbolic x coords={pad0,batch,sample,huf,pad1},
        xmin=pad0, xmax=pad1,
        xtick={batch,sample,huf},
        xticklabels={{Batch\\70\%},{Sample\\54\%},{\textbf{\method{}}\\36\%}},
        xticklabel style={align=center, font=\fontsize{6.5}{7.5}\selectfont},
        enlarge x limits=0,
        label style={font=\scriptsize},
        tick label style={font=\scriptsize},
        axis lines=left,
        axis line style={black!55},
        tick align=outside,
        tick style={black!55},
        ymajorgrids=true,
        grid style={black!10, thin},
        clip=false,
        nodes near coords,
        every node near coord/.append style={
            font=\scriptsize,
            /pgf/number format/fixed,
            /pgf/number format/print sign,
            /pgf/number format/precision=3
        },
    ]
        \addplot[
            ybar, bar width=14pt, bar shift=0pt,
            fill=hufsweep!22, draw=hufsweep, line width=0.8pt,
            every node near coord/.append style={text=hufsweep}
        ] coordinates {(batch,0.074)};

        \addplot[
            ybar, bar width=14pt, bar shift=0pt,
            fill=hufsweep!22, draw=hufsweep, line width=0.8pt,
            every node near coord/.append style={text=hufsweep}
        ] coordinates {(sample,0.036)};

        \addplot[
            ybar, bar width=14pt, bar shift=0pt,
            fill=hufgood!25, draw=hufgood, line width=1.0pt,
            every node near coord/.append style={text=hufgood}
        ] coordinates {(huf,-0.096)};

        \draw[hufbad!70, densely dashed, line width=0.7pt]
            ({rel axis cs:0,0} |- {axis cs:batch,0.211})
            -- ({rel axis cs:1,0} |- {axis cs:batch,0.211});
        \draw[black!55, densely dashed, line width=0.7pt]
            ({rel axis cs:0,0} |- {axis cs:batch,0})
            -- ({rel axis cs:1,0} |- {axis cs:batch,0});

        \node[font=\scriptsize, text=hufbad!85, anchor=south east]
            at ({rel axis cs:1,0} |- {axis cs:batch,0.215}) {all synthetic};
        \node[font=\scriptsize, text=black!55, anchor=north west]
            at ({rel axis cs:0.02,0} |- {axis cs:batch,-0.004}) {real only};

        \node[text=hufgood, font=\small\bfseries, anchor=south]
            at (axis cs:huf,0.055) {$\checkmark$};
    \end{axis}
\end{tikzpicture}
\endgroup

    
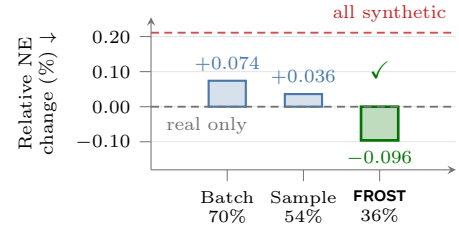
\captionof{figure}{Two-stage ablation. Percentage denotes the synthetic data drop ratio.}
    \label{fig:ads-stage-combination}
\end{minipage}
\par
\endgroup

\section{Conclusion}
\label{sec:conclusion}

We present \method{}, a novel online synthetic-data filtering framework
that selects useful synthetic data to improve downstream task performance. 
\method{} estimates synthetic-data utility through gradient feedback anchored in 
real training data, without requiring an external verifier or a held-out validation
set. 
Experiments on image classification and LLM fine-tuning show that
\method{} improves absolute accuracy with fewer synthetic training samples,
demonstrating robust effectiveness of our approach across different domains, model
architectures, and 3 synthetic image generators. Moreover, in a large-scale
industrial ads re-ranking system, \method{} delivers significant performance
gains over a highly optimized production baseline, flipping the regression from
the full noisy synthetic pool into an improvement over real-only training.

\clearpage
\bibliographystyle{assets/plainnat}
\bibliography{references}

\clearpage
\beginappendix
\section{Theoretical Analysis}
\label{app:theory}

\subsection{Derivation of the Utility Approximation}
\label{app:utility-approximation}

We analyze training benefit as a local real-loss reduction to clarify what
the utility score captures. Fix the current model and real mini-batch
$\mathcal R_t$, and let $\widehat{\mathcal L}_{R,t}(\phi)$ denote its
empirical loss as a function of the scoring block $\phi$, with all other
parameters held fixed. Its gradient is
$\nabla_\phi\widehat{\mathcal L}_{R,t}(\phi_t)=R_t$.
For a hypothetical SGD step
$\phi'=\phi_t-\eta\bar g_{S,t}$, define
\begin{equation}
    \Delta_{R,t}
    =\widehat{\mathcal L}_{R,t}(\phi_t)
     -\widehat{\mathcal L}_{R,t}(\phi').
    \label{eq:local-training-benefit}
\end{equation}
Assuming the empirical real loss is $L$-smooth over this step, Taylor's
theorem gives
\begin{equation}
    \Delta_{R,t}=\eta R_t^\top\bar g_{S,t}+\mathcal H_t,
    \qquad
    |\mathcal H_t|\le
    \frac{L\eta^2}{2}\|\bar g_{S,t}\|^2,
    \label{eq:utility-remainder}
\end{equation}
where $\mathcal H_t$ is the signed higher-order correction. Let
$e_t=R_t-\hat R_t$ be the deviation of the current real-batch gradient from
the EMA reference. Substituting
$R_t=\hat R_t+e_t$ and $U_t=\hat R_t^\top\bar g_{S,t}$ yields
\begin{equation}
    \underbrace{\Delta_{R,t}}_{\text{training benefit}}
    =
    \underbrace{\eta U_t}_{\text{first-order utility}}
    +
    \underbrace{\eta e_t^\top\bar g_{S,t}}_{\text{reference error}}
    +
    \underbrace{\mathcal H_t}_{\text{higher-order effects}},
    \label{eq:utility-decomposition}
\end{equation}
The reference-error and higher-order terms can either increase or decrease
the local benefit. By Cauchy--Schwarz,
\begin{equation}
    \Delta_{R,t}\ge
    \eta U_t
    -\eta\|e_t\|\,\|\bar g_{S,t}\|
    -\frac{L\eta^2}{2}\|\bar g_{S,t}\|^2.
    \label{eq:utility-bound}
\end{equation}
A larger $U_t$ increases the first-order term but does not determine the
other two terms. In particular, large gradient norms can raise the raw
alignment score while also enlarging the higher-order error bound.
Consequently, utility alone need not preserve the ordering of local
training benefits.

Equation~\ref{eq:utility-decomposition} concerns a single update in the
scoring block; the experiment in Section~\ref{sec:utility-horizon} examines subsequent full-model training.
In that comparison, High reaches a higher early accuracy peak but ends the
30-step window with lower mean real-test accuracy than Mid. At this
checkpoint, the highest-utility band is therefore not the best choice for
short-horizon training.

\subsection{Theoretical Guarantee for EMA Reference Tracking}
\label{app:ema-tracking}

In Section~\ref{sec:sample-utility}, we estimate the target real gradient
using a bias-corrected exponential moving average (EMA) of real
mini-batch gradients. Here, we show that the expected squared error
between the EMA reference $\hat R_t$ and the population real gradient
$\bar g_t=\nabla_\phi\mathcal L_R(\theta_t)$ in the scoring block $\phi$
is bounded under the assumptions below.

\noindent\textbf{Assumptions.}
Let $\bar g_t=\nabla_\phi\mathcal L_R(\theta_t)$ denote the population
real gradient at step $t$. We assume the following conditions for the
real-data gradients and optimization process:
\begin{enumerate}
    \item \textbf{Unbiased mini-batch gradients.}
    Let $\mathcal F_{k-1}$ denote the complete training history before
    sampling the real mini-batch at step $k$, including the current
    parameters $\theta_k$. Thus, $\theta_k$ and $\bar g_k$ are
    $\mathcal F_{k-1}$-measurable. We assume
    \begin{equation}
        \mathbb E[R_k-\bar g_k\mid\mathcal F_{k-1}]=0.
        \label{eq:ema-noise-assumptions}
    \end{equation}
    \item \textbf{Bounded conditional variance.}
    The conditional variance of the real mini-batch gradient noise
    is bounded by $\nu_R^2$:
    $\mathbb E[\|R_k-\bar g_k\|^2\mid\mathcal F_{k-1}]
    \le\nu_R^2$ almost surely.
    \item \textbf{$L$-smoothness.} The population real-data objective is
    $L$-smooth in the full parameter vector:
    $\|\nabla_\theta\mathcal L_R(\theta_x)
    -\nabla_\theta\mathcal L_R(\theta_y)\|
    \le L\|\theta_x-\theta_y\|$.
    \item \textbf{Bounded parameter updates.} The actual full-model
    update satisfies $\|\theta_{k+1}-\theta_k\|\le S$ almost surely,
    where $S$ is a uniform bound on the update norm.
\end{enumerate}

\noindent\textbf{Result.}
Under the assumptions above, the expected squared tracking error of the
bias-corrected EMA in Equation~\ref{eq:real-gradient-ema} is bounded by
the following expression for fixed $0\le\beta<1$ and $t\ge1$:
\begin{equation}
    \mathbb E\|\hat R_t-\bar g_t\|^2
    \le 2\nu_R^2\left(\frac{1-\beta}{1+\beta}\right)
            \frac{1}{(1-\beta^t)^2}
       +2L^2S^2\left(\frac{\beta}{1-\beta}\right)^2
            \frac{1}{(1-\beta^t)^2}.
    \label{eq:ema-tracking-bound}
\end{equation}

\noindent\textbf{Proof.}
From Equation~\ref{eq:real-gradient-ema}, the bias-corrected EMA can be
unrolled as a weighted sum of past mini-batch gradients:
\begin{equation}
    \hat R_t=\sum_{k=1}^t w_{t,k}R_k,
    \qquad
    w_{t,k}=\frac{(1-\beta)\beta^{t-k}}{1-\beta^t},
    \label{eq:ema-history-weights}
\end{equation}
The weights sum to one, $\sum_{k=1}^t w_{t,k}=1$. We decompose the
error into mini-batch noise and tracking lag from parameter drift:
\[
    \hat R_t-\bar g_t
    =\underbrace{\sum_{k=1}^t w_{t,k}(R_k-\bar g_k)}_{\text{Noise}}
     +\underbrace{\sum_{k=1}^t w_{t,k}(\bar g_k-\bar g_t)}_{\text{Tracking Lag}}.
\]
Using $\|a+b\|^2\le2\|a\|^2+2\|b\|^2$ and taking expectations gives
\[
    \begin{aligned}
        \mathbb E\|\hat R_t-\bar g_t\|^2
        \le{}&2\,\mathbb E\left\|
            \sum_{k=1}^t w_{t,k}(R_k-\bar g_k)\right\|^2\\
        &+2\,\mathbb E\left\|
            \sum_{k=1}^t w_{t,k}(\bar g_k-\bar g_t)\right\|^2.
    \end{aligned}
\]

\noindent\textbf{1. Bounding the Noise Term.}
The mini-batch gradient noise has zero conditional mean.
For $j<k$, the earlier noise $R_j-\bar g_j$ is
$\mathcal F_{k-1}$-measurable. By the tower property,
\[
    \begin{aligned}
        \mathbb E\langle R_j-\bar g_j,R_k-\bar g_k\rangle
        &=\mathbb E\!\left[
            \left\langle R_j-\bar g_j,
            \mathbb E[R_k-\bar g_k\mid\mathcal F_{k-1}]
            \right\rangle\right]\\
        &=0.
    \end{aligned}
\]
Thus, the cross terms vanish, giving
\[
    \mathbb E\left\|\sum_{k=1}^t w_{t,k}(R_k-\bar g_k)\right\|^2
    =\sum_{k=1}^t w_{t,k}^2\mathbb E\|R_k-\bar g_k\|^2
    \le\nu_R^2\sum_{k=1}^t w_{t,k}^2.
\]
Bounding the sum of squared weights by an infinite geometric series,
\begin{equation}
    \begin{aligned}
        \sum_{k=1}^t w_{t,k}^2
        &=\frac{(1-\beta)^2}{(1-\beta^t)^2}
            \sum_{k=1}^t\beta^{2(t-k)}\\
        &\le\frac{(1-\beta)^2}{(1-\beta^t)^2}
            \sum_{a=0}^{\infty}\beta^{2a}
         =\frac{1-\beta}{(1-\beta^t)^2(1+\beta)}.
    \end{aligned}
    \label{eq:ema-weight-sums}
\end{equation}
Thus, the noise contribution is bounded by
$\nu_R^2(1-\beta)/[(1-\beta^t)^2(1+\beta)]$.

\noindent\textbf{2. Bounding the Tracking Lag Term.}
By $L$-smoothness and the bounded-update assumption, the population
gradient drift in the fixed scoring block satisfies
\[
    \|\bar g_k-\bar g_t\|
    \le L\|\theta_k-\theta_t\|
    \le L\sum_{j=k}^{t-1}\|\theta_{j+1}-\theta_j\|
    \le LS(t-k).
\]
Applying this bound to the weighted sum gives, almost surely,
\[
    \begin{aligned}
        \left\|\sum_{k=1}^t w_{t,k}(\bar g_k-\bar g_t)\right\|
        &\le\sum_{k=1}^t w_{t,k}\|\bar g_k-\bar g_t\|\\
        &\le\frac{LS(1-\beta)}{1-\beta^t}
            \sum_{k=1}^t\beta^{t-k}(t-k).
    \end{aligned}
\]
With $a=t-k$, the arithmetico-geometric series yields
\[
    \sum_{k=1}^t\beta^{t-k}(t-k)
    \le\sum_{a=0}^{\infty}a\beta^a
    =\frac{\beta}{(1-\beta)^2}.
\]
Therefore, the tracking lag is bounded by
$LS\beta/[(1-\beta^t)(1-\beta)]$. Squaring this bound and combining
it with the noise bound proves
Equation~\ref{eq:ema-tracking-bound}.

Equation~\ref{eq:ema-tracking-bound} establishes a bias--variance
trade-off for the EMA reference. As $t$ grows, the $(1-\beta^t)$ terms
approach one. In this long-history regime, a high $\beta$, such as
$0.99$, suppresses the mini-batch noise contribution, while introducing
a tracking-lag term proportional
to $L^2S^2[\beta/(1-\beta)]^2$. Small parameter updates control this
lag, supporting the use of EMA when the real-gradient trajectory changes
slowly.

\section{Implementation Details}
\label{app:implementation-details}

\noindent\textbf{Filtering configuration.}
We use a rolling-buffer window of $W=200$ batch scores for image
classification and ads re-ranking, and $W=50$ for LLM fine-tuning.

\subsection{Image classification}
\label{app:image-implementation}

\noindent\textbf{Data and training.}
CIFAR-100 \citep{krizhevsky2009learning} contains 100 classes, with 500
training and 100 test images per class. We use a CIFAR-style ResNet-18
\citep{he2016resnet} with $[2,2,2,2]$ BasicBlocks across four stages of
widths $64/128/256/512$, a $3\times3$ convolutional stem with stride $1$,
and a linear $512\to100$ classifier. We use SGD with momentum $0.9$ and
weight decay $5\times10^{-4}$. We train for 200 epochs with a batch
size of 1,024 and a learning rate of $0.2$ with a OneCycle schedule,
using real and synthetic images. We evaluate accuracy on the 10,000 real
test images.

\begin{figure}[!htb]
    \centering
    \includegraphics[width=0.98\linewidth]{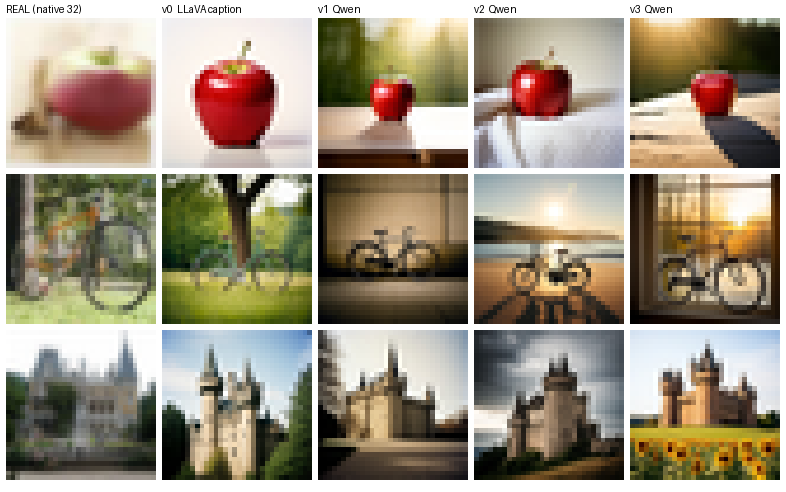}
    \par\smallskip
    {\small
    \begin{tabular}{@{}p{0.24\linewidth}p{0.72\linewidth}@{}}
        \textbf{Bicycle v0 (excerpt)} &
        a bicycle leaning against a tree in a grassy area. \\
        \textbf{Bicycle v2 (Qwen)} &
        a bicycle gleaming on a sandy beach, with the sun setting behind a range of hills.
    \end{tabular}
    }
    \caption{Examples from the image-generation pipeline. Rows show apple, bicycle, and castle. The first column contains real CIFAR-100 images; the remaining columns are SANA-1.6B outputs conditioned on the cleaned LLaVA caption (v0) and three Qwen rewrites (v1--v3). Generated images are downsampled to $32\times32$ with Lanczos and enlarged with nearest-neighbor interpolation for display; real images are enlarged from their native $32\times32$ resolution. The bicycle captions illustrate how rewriting changes the scene while preserving the subject.}
    \label{fig:cifar100-generation}
\end{figure}

\noindent\textbf{Generation pipeline.}
We construct the FLUX.1~[schnell] and SANA pools through image captioning,
prompt diversification, and text-to-image generation. Captioning and
rewriting are performed once offline, cached as JSONL, and shared by
both generators. Figure~\ref{fig:cifar100-generation} shows examples of
the real images and the resulting synthetic variants.

\noindent\textbf{Image captioning.}
We use LLaVA-NeXT-Mistral-7B \citep{liu2024llavanext}, specifically
\texttt{llava-hf/llava-v1.6-mistral-7b-hf}, to caption each of the 50,000
real training images. Images are bicubically upsampled from $32\times32$
to $336\times336$. We use greedy decoding in bfloat16, a batch size of
16, and at most 150 new tokens. The prompt supplies the class name and
asks for the setting, colors, viewpoint, and lighting, while excluding
references to image quality or resolution. A deterministic cleaning pass
removes residual quality descriptors and opening boilerplate, trims
incomplete trailing sentences, and normalizes whitespace. We prepend
the class name to the cleaned caption, disambiguating class names when
needed, and retain this base prompt as v0.

\noindent\textbf{Prompt diversification.}
Qwen2.5-7B-Instruct \citep{qwen2024qwen25} generates three rewrites
(v1--v3) of each cleaned caption, using temperature $0.9$, top-$p$ $0.95$,
a batch size of 32, and at most 400 new tokens. We request a JSON array
of three distinct, single-sentence captions under 25 words each. The
instructions preserve the subject class while varying the background,
surface, lighting, time of day, and camera angle. They also request one
large, centered subject with a simple background and prohibit wide
shots, multiple subjects, and image-quality descriptors. These framing
constraints aim to keep the labeled object recognizable after
downsampling to the classifier's $32\times32$ input resolution.

\noindent\textbf{Text-to-image generation.}
FLUX.1~[schnell] \citep{blackforestlabs2024flux1schnell} and SANA-1.6B
\citep{xie2025sana} each generate one image per conditioning string:
the base caption v0 and three Qwen variants v1--v3. Each generator thus
produces four synthetic examples per real image, giving 2,000 images per
class and 200,000 images in total. Separately, we use a Stable Diffusion
v1.4 pool
\citep{rombach2022ldm} from the dataset release of
\citet{shipard2023diversity}, containing 1,800 images per class.
This released pool is not generated through our caption pipeline.

\noindent\textbf{Image features and reference data.}
For the feature-based image selection baselines, we use the final
pooled-and-projected 512-dimensional image embeddings from a frozen
OpenAI CLIP ViT-B/32 encoder \citep{radford2021clip}. In our
feature-extraction pipeline, images are bicubically resized to
$224\times224$, normalized using ImageNet statistics, and processed
without data augmentation. This normalization follows the authors' public
implementation for DS3 and CovMatch \citep{rezaei2026highdimensional},
which uses ImageNet statistics instead of CLIP's default preprocessing.
Embeddings are cached in FP16, then converted
to FP32 and $\ell_2$-normalized before selection. Selection is performed
independently for each CIFAR-100 class, using all 500 real training images
of that class as references.

\noindent\textbf{Covariance Matching.}
For CovMatch \citep{rezaei2026highdimensional}, we fit a 32-dimensional
principal component analysis (PCA) projection on each class's real
features and apply the same projection to its real and synthetic
features. We greedily select synthetic examples to minimize the
Frobenius distance between the unbiased sample covariance of the
selected synthetic features and the corresponding real-data covariance.

\noindent\textbf{DS3.}
We implement the nearest-cluster selection strategy from DataS$^3$
\citep{hulkund2025datas3}, adapted to synthetic pools by
\citet{rezaei2026highdimensional}. For each class, we run $k$-means on
the synthetic features, starting with $K=200$ clusters and using
$k$-means++ initialization \citep{arthur2007kmeanspp} and Lloyd's algorithm
\citep{lloyd1982least}. A cluster is eligible if its centroid is the
nearest centroid to at least one real reference example. We uniformly
sample the exact selection budget from the union of eligible clusters.
If this union contains too few examples, we successively halve $K$ and
repeat clustering and eligibility selection until the budget can be met.

\noindent\textbf{Prompt templates.}
The following templates are used for captioning and diversification.
\texttt{\{class\_name\}} denotes the subject class,
\texttt{\{caption\}} the cleaned LLaVA caption, and
\texttt{\{n\}} is set to 3. LLaVA uses the Mistral instruction format;
Qwen receives separate system and user messages.

\Needspace{9\baselineskip}
\begin{lstlisting}[style=generationprompt,title={\textbf{LLaVA-NeXT: captioning prompt}}]
[INST] <image>
This is a photo of a {class_name}. In one sentence, describe the scene: the
setting or background, the main colours, the camera viewpoint, and the
lighting. Describe it as a normal photograph. Do not mention image quality,
resolution, pixelation, blurriness, or that it is small or a thumbnail. Do
not begin with 'The image' or 'This image'. [/INST]
\end{lstlisting}

\begin{lstlisting}[style=generationprompt,title={\textbf{Qwen2.5-7B-Instruct: system prompt}}]
You rewrite image captions into variants for a text-to-image generator. Every
variant keeps one single subject large and centred in the frame. You reply with
JSON and nothing else.
\end{lstlisting}

\begin{lstlisting}[style=generationprompt,title={\textbf{Qwen2.5-7B-Instruct: user prompt}}]
Subject: "{class_name}"
Base caption: "{caption}"

Write exactly {n} DIFFERENT one-sentence captions of the SAME subject.

Every caption must:
- show ONE single {class_name}, close up, filling most of the frame;
- vary the background, surface, lighting, time of day and camera angle
  relative to the base caption and to each other;
- keep the background simple and secondary -- a short phrase at most;
- stay photographic and physically plausible;
- be under 25 words.

Never do any of these:
- zoom out, or describe a wide shot, landscape, skyline, street scene or room;
- show several {class_name}s, a crowd, a pile, a collage or a grid;
- make something other than the {class_name} the main subject;
- mention image quality, resolution, pixelation or blurriness.

Reply with only a JSON array of {n} strings.
\end{lstlisting}
\par

\subsection{LLM fine-tuning for text-to-SQL}
\label{app:llm-implementation}

\noindent\textbf{Model and input format.}
We fully fine-tune Qwen2.5-Coder-0.5B-Instruct \citep{hui2024qwen25coder} 
in bfloat16, without adapters. We use the model's chat template with 
a fixed system prompt. The user turn contains the database's 
\texttt{CREATE TABLE} schema, external knowledge when
available, and the question, followed by \texttt{SQLite query:}. Schemas
are read directly from SQLite files using the same serialization for
training and inference. The target is whitespace-normalized gold SQL
followed by EOS. We mask prompt tokens from the loss and left-truncate
the prompt when necessary to preserve the target within 2,048 tokens.

\noindent\textbf{Optimization and inference.}
We train for one epoch with a global batch size of 64 and AdamW
($\beta_1=0.9$, $\beta_2=0.95$, $\epsilon=10^{-8}$, zero weight decay).
The learning rate is $5\times10^{-5}$, with a $10\%$ warmup
followed by cosine decay to zero. Gradients are clipped to a global
$\ell_2$ norm of $1.0$ after accumulation. At inference, we use greedy
decoding with at most 160 new tokens and report EX on Spider~1.0 dev.

\noindent\textbf{Training objective.}
We average cross-entropy over each example's supervised tokens, then
average over retained examples in the global batch. At step $t$, let
$\mathcal T_i$ be example $i$'s supervised token positions, $z_{ik,t}$
its logits, and $y_{ik}$ the target token. The objective is
\begin{equation}
    \ell_{i,t}=\frac{1}{|\mathcal T_i|}
        \sum_{k\in\mathcal T_i}\mathrm{CE}(z_{ik,t},y_{ik}),
    \qquad
    \mathcal L_t=
        \frac{\sum_{i\in\mathcal G_t}w_i\ell_{i,t}}
             {\sum_{i\in\mathcal G_t}w_i},
    \label{eq:llm-training-objective}
\end{equation}
where $\mathcal G_t$ is the global batch and $w_i\in\{0,1\}$ is the
filter's keep weight; real examples always have $w_i=1$. The denominator
includes all accumulation micro-batches. Normalizing by the kept count
avoids shrinking the loss by the retained fraction. Cross-entropy is
computed in FP32.

\noindent\textbf{Token-wise rank-1 gradient computation.}
We compute utility without constructing a full LM-head gradient for each
example. For scoring, we use the output-projection gradient with hidden
states held fixed, excluding the input-embedding path when weights are
tied. Let $f_{ik,t}\in\mathbb R^d$ be the hidden state and
$\delta_{ik,t}=\operatorname{softmax}(z_{ik,t})-\operatorname{onehot}(y_{ik})
\in\mathbb R^V$ the gradient of token-level cross-entropy with respect
to the logits. Each token contributes a rank-1
outer product, so the example's head gradient is
\begin{equation}
    G_{i,t}=\frac{1}{|\mathcal T_i|}
        \sum_{k\in\mathcal T_i}\delta_{ik,t}f_{ik,t}^{\top},
    \qquad g_{i,t}=\operatorname{vec}(G_{i,t}).
    \label{eq:llm-head-gradient}
\end{equation}
Let $\widehat{\mathbf R}_t\in\mathbb R^{V\times d}$ be the matrix form
of the bias-corrected reference $\hat R_t$ in
Equation~\ref{eq:real-gradient-ema}. Reordering the inner product gives
\begin{equation}
    u_{i,t}
    =\langle\widehat{\mathbf R}_t,G_{i,t}\rangle_F
    =\frac{1}{|\mathcal T_i|}\sum_{k\in\mathcal T_i}
        \delta_{ik,t}^{\top}
        \bigl(\widehat{\mathbf R}_t f_{ik,t}\bigr).
    \label{eq:llm-factorized-utility}
\end{equation}
This computes the same head-level dot product without materializing
$G_{i,t}$. With $V=151{,}936$ and $d=896$, a single FP32 head-gradient
matrix would occupy approximately $0.51$~GiB, or over $32$~GiB for 64
examples. We instead process supervised tokens in chunks of at most 256,
bounding token-wise temporary storage while retaining one shared
real-reference matrix.

\subsection{Computational overhead}
\label{app:computational-overhead}

\noindent\textbf{Utility computation.}
We retain the sample utility $u_{i,t}=\hat R_t^\top g_{i,t}$ from
Equation~\ref{eq:sample-utility}. Here $m=|S_t|$ is the number of
synthetic candidates before filtering, and $|\mathcal R_t|$ is the
number of real examples in the same training step.
Bias terms are included in the image
utility scores but omitted from the leading-order cost analysis below.
For a linear prediction head with input dimension $d$ and output
dimension $C$, let $f_{i,t}\in\mathbb R^d$ be the input feature and
$\delta_{i,t}=\operatorname{softmax}(z_{i,t})-\operatorname{onehot}(y_i)
\in\mathbb R^C$ the cross-entropy gradient with respect to the logits,
where $z_{i,t}$ and $y_i$ are the example's logits and class label.
Following the matrix notation in Appendix~\ref{app:llm-implementation},
write the weight gradient as $G_{i,t}=\delta_{i,t}f_{i,t}^{\top}$ and
the corresponding weight component of the same EMA reference $\hat R_t$
as $\widehat{\mathbf R}_t\in\mathbb R^{C\times d}$.
Its contribution to the utility is
\begin{equation}
    \langle\widehat{\mathbf R}_t,G_{i,t}\rangle_F
    =\delta_{i,t}^{\top}
      \bigl(\widehat{\mathbf R}_t f_{i,t}\bigr).
    \label{eq:head-utility-cost}
\end{equation}
Given the features and logit gradients, scoring $m$ synthetic examples costs
$O(mCd)$ without materializing per-example head-gradient matrices.
Aggregating the reference from $\mathcal R_t$ costs
$O(|\mathcal R_t|Cd)$, and updating the EMA costs $O(Cd)$.
The head-level computation uses
$O(Cd+(|\mathcal R_t|+m)(C+d))$ storage for the reference, features,
logit gradients, and intermediate products, avoiding an
$O((|\mathcal R_t|+m)Cd)$ per-example gradient tensor.
The full model is still trained; these costs concern only head-level
scoring.
For LLMs, the analogous computation is performed per supervised token,
using the factorization and chunking described in
Appendix~\ref{app:llm-implementation}.

Batch aggregation and calibration cost $O(m+W)$ when buffer statistics
are computed directly. Sorting-based IQR filtering costs $O(m\log m)$
and is needed only for out-of-band batches; utility scores are computed
for all synthetic candidates before routing.

\noindent\textbf{ResNet-18 arithmetic cost.}
We compare head-level scoring with full-model training using the
convolutional structure of ResNet-18. We measure arithmetic cost in
multiply--accumulate operations (MACs), counting one multiplication and
accumulation as one MAC. For convolution $j$, let $H_j,W_j$ denote its output
height and width, $c_j^{\mathrm{in}},c_j^{\mathrm{out}}$ its input and
output channel counts, and $k_j$ its kernel width. Its forward cost is
$H_jW_jc_j^{\mathrm{in}}c_j^{\mathrm{out}}k_j^2$ MACs per image.
Summing over the backbone, including the stem and projection shortcuts,
gives
\[
    F_{\mathrm{backbone}}
    =\sum_j H_jW_jc_j^{\mathrm{in}}c_j^{\mathrm{out}}k_j^2.
\]
The classifier adds $Cd$ MACs. Backpropagation computes both weight and
activation gradients, with a dominant convolution/linear cost of
approximately twice the forward cost. Table~\ref{tab:resnet-macs}
summarizes these terms for a batch of $|\mathcal R_t|+m$ examples.

\begin{table}[H]
    \caption{Leading arithmetic costs for regular ResNet-18 training and head-level utility computation. The latter assumes that features and logit gradients are already available.}
    \label{tab:resnet-macs}
    \centering
    \small
    \begin{tabular}{ll}
        \toprule
        Computation & MACs per step \\
        \midrule
        Regular forward & $(|\mathcal R_t|+m)(F_{\mathrm{backbone}}+Cd)$ \\
        Regular backward & $\approx2(|\mathcal R_t|+m)(F_{\mathrm{backbone}}+Cd)$ \\
        \midrule
        Real-reference aggregation & $|\mathcal R_t|Cd$ \\
        Synthetic utility scoring & $mCd$ \\
        EMA update & $O(Cd)$ \\
        \bottomrule
    \end{tabular}
\end{table}

Thus, the estimated regular forward/backward cost is
\begin{equation}
    \mathrm{MAC}_{\mathrm{train}}
    \approx3(|\mathcal R_t|+m)(F_{\mathrm{backbone}}+Cd),
    \label{eq:resnet-training-macs}
\end{equation}
whereas the additional reference aggregation and scoring matrix
operations require
\begin{equation}
    \mathrm{MAC}_{\mathrm{head}}=(|\mathcal R_t|+m)Cd,
    \label{eq:resnet-scoring-macs}
\end{equation}
apart from the EMA update and lower-order score operations.
The gap arises because backbone convolutions operate across spatial
positions and multiple layers, with cost quadratic in channel width
for equal-width convolutions, while head scoring operates on pooled
features with cost linear in $d$ for fixed $C$.
Specifically, let $A_d$ be the spatial area of the final ResNet stage
and $N_d$ the number of its $d\to d$ convolutions with kernel width $k$.
These convolutions alone contribute $N_dA_dk^2d^2$ MACs per image, so
$F_{\mathrm{backbone}}\ge N_dA_dk^2d^2$. Under the above cost model,
\begin{equation}
    \frac{\mathrm{MAC}_{\mathrm{head}}}{\mathrm{MAC}_{\mathrm{train}}}
    \approx\frac{Cd}{3(F_{\mathrm{backbone}}+Cd)}
    \le\frac{C}{3N_dA_dk^2d}.
    \label{eq:resnet-mac-ratio}
\end{equation}
For the CIFAR-style ResNet-18, $N_dA_dk^2d\gg C$, making the head-level
matrix computation much smaller than regular forward/backward training.
This comparison concerns convolution/linear arithmetic rather than
end-to-end runtime.

\noindent\textbf{Measured training throughput.}
Because the added arithmetic is negligible, the measured overhead is
dominated by memory traffic and synchronization rather than by FLOPs.
Compared with regular training, \method{} reduces training throughput
by approximately $6.8\%$ in image classification and $5.7\%$ in
large-scale industrial ads re-ranking, where utility is computed using
the final shared-backbone layer. Unlike MAC counts, end-to-end throughput
also reflects implementation overheads, which can include logging,
communication, and CPU--GPU synchronization. These measurements indicate
modest overhead in both training settings.

\section{Additional Results}
\label{app:additional-results}

\subsection{Two-stage ablation on Spider~1.0, by difficulty}
\label{app:llm-ablation-table}

Table~\ref{tab:llm-ablation} gives the per-difficulty breakdown behind
Figure~\ref{fig:llm-stage-combination}. Full \method{} improves overall EX
over Batch gate only and Sample IQR only by $5.0$ and $2.6$ percentage
points while dropping less synthetic data than either. On Extra Hard,
the gains are $9.1$ and $6.7$ points, respectively. Sample IQR only
performs better than full \method{} on Hard by $1.7$ points, but has
lower EX in the other three difficulty categories.

\begin{table}[H]
    \caption{Two-stage ablation on Spider~1.0 dev. Full \method{} is the reference. \textit{Drop} is the synthetic-data drop ratio (\%); EX is execution accuracy (\%). $\Delta$EX is the overall EX change relative to full \method{}, in percentage points (pp). All variants retain the real training data.}
    \label{tab:llm-ablation}
    \centering
    \small
    \setlength{\tabcolsep}{3pt}
    \begin{tabular*}{\linewidth}{@{\extracolsep{\fill}}lccccccc@{}}
        \toprule
        & & & & \multicolumn{4}{c}{EX by difficulty (\%) $\uparrow$} \\
        \cmidrule(lr){5-8}
        Variant & \textit{Drop} & EX $\uparrow$ & $\Delta$EX $\uparrow$ & Easy & Medium & Hard & Extra Hard \\
        \midrule
        Full \method{} & \textit{20.4} & \textbf{65.7} & $0.0$ & 83.5 & 68.6 & 54.6 & 42.8 \\
        \midrule
        Batch gate only & \textit{80.0} & 60.7 & $-5.0$ & 79.0 & 63.5 & 53.4 & 33.7 \\
        Sample IQR only & \textit{32.2} & 63.1 & $-2.6$ & 80.6 & 65.9 & 56.3 & 36.1 \\
        \bottomrule
    \end{tabular*}
\end{table}

\end{document}